# 3D AGSE-VNet: An Automatic Brain Tumor MRI Data Segmentation Framework

Xi Guan[1,*], Guang Yang[2,3,*,†], Jianming Ye[4], Weiji Yang[5], Xiaomei Xu[1], Weiwei Jiang[1], Xiaobo Lai [1,†]

[1] School of Medical Technology and Information Engineering, Zhejiang Chinese Medical University, Hangzhou, 310053, China
[2] Cardiovascular Research Centre, Royal Brompton Hospital, London, SW3 6NP, UK
[3] National Heart and Lung Institute, Imperial College London, London, SW7 2AZ, UK
[4] First Affiliated Hospital, Gannan Medical University, Ganzhou, 341000, China
[5] College of Life Science, Zhejiang Chinese Medical University, Hangzhou, 310053, China

[*] Co-first authors contributed equally.
[†] Corresponding Authors: G. Yang (g.yang@imperial.ac.uk) and X. Lai (dmia_lab@zcmu.edu.cn)

**Abstract**

Background Glioma is the most common brain malignant tumor, with a high morbidity rate and a mortality rate of more than three percent, which seriously endangers human health. The main method of acquiring brain tumors in the clinic is MRI. Segmentation of brain tumor regions from multi-modal MRI scan images is helpful for treatment inspection, post-diagnosis monitoring, and effect evaluation of patients. However, the common operation in clinical brain tumor segmentation is still manual segmentation, lead to its time-consuming and large performance difference between different operators, a consistent and accurate automatic segmentation method is urgently needed. With the continuous development of deep learning, researchers have designed many automatic segmentation algorithms; however, there are still some problems: 1) The research of segmentation algorithm mostly stays on the 2D plane, this will reduce the accuracy of 3D image feature extraction to a certain extent. 2) MRI images have gray-scale offset fields that make it difficult to divide the contours accurately.

Methods To meet the above challenges, we propose an automatic brain tumor MRI data segmentation framework which is called AGSE-VNet. In our study, the Squeeze and Excite (SE) module is added to each encoder, the Attention Guide Filter (AG) module is added to each decoder, using the channel relationship to automatically enhance the useful information in the channel to suppress the useless information, and use the attention mechanism to guide the edge information and remove the influence of irrelevant information such as noise.

Results We used the BraTS2020 challenge online verification tool to evaluate our approach. The focus of verification is that the Dice scores of the whole tumor (WT), tumor core (TC) and enhanced tumor (ET) are 0.68, 0.85 and 0.70, respectively.

Conclusion Although MRI images have different intensities, AGSE-VNet is not affected by the size of the tumor, and can more accurately extract the features of the three regions, it has achieved impressive results and made outstanding contributions to the clinical diagnosis and treatment of brain tumor patients.

# 1. Introduction

Glioma is one of the common types of primary brain tumors, accounting for about 50% of intracranial tumors[1]. According to the WHO classification criteria, gliomas can be divided into four grades according to different symptoms, of which I and II are low-grade gliomas (LGG), III and IV are high-grade gliomas (HGG) [2]. Due to the high mortality rate of glioma, it can appear in any part of the brain and people of any age, with various histological subregions and varying degrees of invasiveness [3]. Therefore, it has attracted widespread attention in the medical field. Because glioblastoma (GBM) cells are immersed in the healthy brain parenchyma and infiltrate the surrounding tissues, they can grow and spread rapidly near the protein fibers, and the deterioration process is very rapid. Therefore, early diagnosis and treatment are essential.

At present, the methods of acquiring brain tumors in clinical practice are mainly computed tomography (CT), positron emission tomography (PET), and magnetic resonance imaging (MRI) [4]. Among them, MRI has become the preferred medical imaging method for brain diagnosis and treatment planning. Because it provides images with high-contrast soft tissue and high spatial resolution [5], it is a good representation of the anatomical structure of the cranial nerve soft tissue and the image of the lesion. At the same time, MRI images can obtain multiple sequence information of brain tumors in different spaces through one scan. This information includes four sequences of T1 weighting (T1), T1-weighted contrast-enhanced (T1-CE), and T2 weighting (T2), fluid attenuation inversion recovery (FLAIR) [6-7]. However, manually segmenting tumors from MRI images requires professional prior knowledge, which is time-consuming and labor-intensive, and is prone to errors, which is very dependent on the doctor's experience. Therefore, the development of an accurate, reliable, and fully automatic brain tumor segmentation algorithm has strong clinical significance.

With the development of computer vision and pattern recognition, convolutional neural networks have been implemented to solve many challenging tasks. For example, classification, segmentation and target detection capabilities have been greatly improved. In addition, deep learning technology shows great potential in medical image processing. So far, plenty of research studies on medical image segmentation have been developed in both academia and industry. VNet [8] has good segmentation performance in single-modal images, but there are still some shortcomings for multi-modal segmentation. In this article, inspired by the integration of the "Project and Excite" (PE) module into the 3D U-net proposed by Anne-Marie et al [9] , we proposed an automatic brain tumor MRI Data segmentation framework, which is called 3D AGSE-VNet. The network structure is shown in Fig. 1. The main contributions of this paper are: 1) Propose a combined segmentation model based on VNet, integrating SE module and AG module. 2) Using volume input, three-dimensional convolution is used to process MRI images. 3) Get excellent segmentation results, have the potential clinical application.

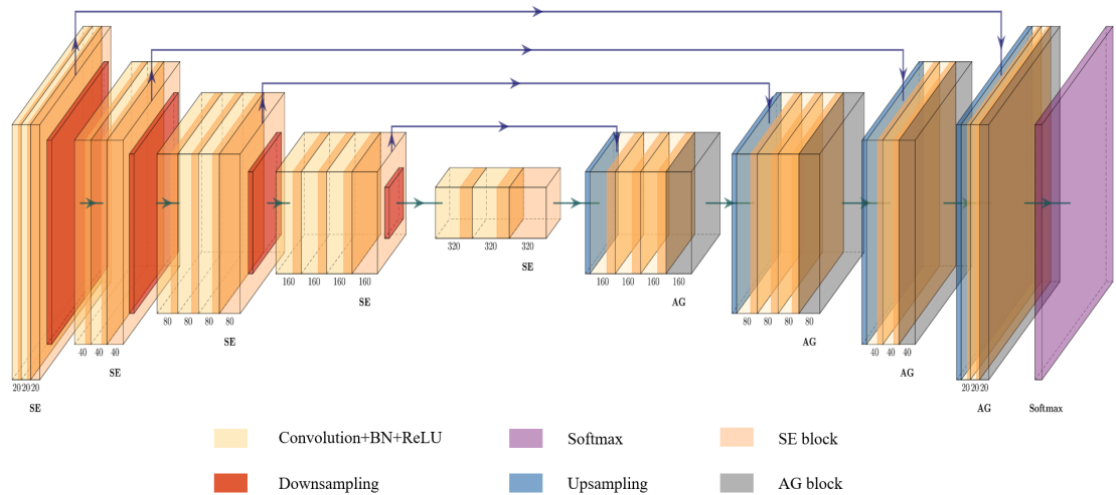

**Fig. 1.** The overall architecture of the proposed 3D AGSE-VNet.

## 2. Related Works

### 2.1. Traditional machine learning

At present, in clinical medicine, it is the goal that experts and scholars have been pursuing to use fully automatic segmentation methods to replace tedious manual segmentation or semi-automatic segmentation. It is also the focus and key technology of medical impact research in recent years. Traditional image processing brain tumor segmentation algorithms use threshold-based segmentation methods, region-based segmentation methods, and boundary-based segmentation methods. Image segmentation based on the threshold is one of the simplest and most traditional methods in image processing. Tustison et al proposed a two-stage segmentation framework based on Random Forest-derived probabilities, using the output of the first classifier to improve the segmentation result of the second classifier [10]. Stadlbauer et al [11] proposed using the normal distribution of data to obtain the threshold. According to the intensity change of each region, an adaptive threshold segmentation method was proposed to separate the foreground image from the background. However, this method has high limitations, and segmentation fails when multiple organizational structures overlap. Amiri et al [12] proposed a multi-layer structure in which structural Random Forest and Bayesian networks are embedded to learn tumor features better, but inputting a large number of features can easily lead to dimensional disasters and waste plenty of time [13] uses a seed region growth algorithm to process brain MRI images according to the threshold *T* and the generation of PD images, and then uses the Markov logic algorithm to process them further to improve segmentation performance.

*2.2. Deep learning*

In recent years, convolutional neural networks have become the most popular method in image classification, and they are widely used in medical image analysis. Sérgio Pereira et al. [14] proposed an automatic positioning method based on Convolutional Neural Network (CNN) to explore the 3×3 core size, use a small core to design a deeper architecture, and use intensity normalization is used as a preprocessing step to train the core validation data set in BraTS 2015. In the article by Hao et al., a fully automatic brain tumor segmentation method based on U-net deep convolutional network was proposed and evaluated on the BraTS 2015 dataset. Cross-validation shows that it can effectively obtain promising segmentation [15]. Wang et al. proposed a cascade network. The first step is to segment the entire tumor. The second step is to segment the tumor nucleus using the obtained bounding box and segment the enhanced tumor nucleus according to the bounding box of the tumor nucleus segmentation result. Use anisotropic convolution and unfolded convolution, combined with multi-view fusion methods to reduce false positives [16]. Andriy Myronenko proposed a 3D MRI tumor subregion segmentation semantic network based on the encoder-decoder structure, which uses auto-encoder branches to reconstruct images, and won first place in the 2018 BraTS Challenge [17]. Feng Xue proposed an integrated 3D U-net brain tumor segmentation method. Using an integrated modelling method, the encoder and decoder are input into 6 networks with different colour block sizes and loss weights, and training has improved various performances [18]. In 2019, Nabilibtehaz et al. developed a novel architecture based on u-net, multires-unet, which increased the extension of residual connections and proposed the residual path (respath). It has verified its use in ISIC and BraTS. Good segmentation performance on the dataset [19]. Xu et al. proposed progressive sequential causality to synthesize high-quality LGR-equivalent images and accurately segment all tissues related to the diagnosis to obtain highly accurate diagnostic indicators in a real clinical environment [20]. Zhou et al. proposed an effective 3D residual neural network for brain tumor segmentation, using a computationally efficient network 3D shuffleNetV2 as an encoder, and introducing a decoder with residual blocks to achieve high-efficiency segmentation [21]. Saman et al. proposed an active contour model driven by optimized energy function for MR brain tumor segmentation with uneven intensity correction and a method to identify and segment brain tumor slices in MRI images [22]. Liu et al. studied a deep learning model based on learnable group convolution and deep supervision. This method replaces the convolution in the feature extraction stage with learnable group convolution. Tests on the BraTS2018 dataset show that the segmentation effect on the core area of the tumor is perfect, surpassing the winning method NVDLMED [23]. In addition, CNN has also been widely used in other medical image analysis tasks. For example, Yurttakal et al. used the convolutional neural network method for laryngeal histopathological image segmentation, which is of great help to the early detection, monitoring and treatment of laryngeal cancer, and rapid and accurate tumor segmentation [24].

*2.3. Our work*

Although many experts and scholars have proposed a variety of deep learning network structures, and have achieved good results in the field of brain tumor segmentation. However, due to the inherent anisotropy of brain glial tumors, MRI images show a high degree of non-uniformity and irregular shapes [25]. Secondly, the segmentation method of deep learning requires large-scale annotation data, while brain tumor data is generally small and complex, and its inherent high heterogeneity will cause intra-class differences between the sub-regions of the brain tumor area and the tumor area, the difference between classes and non-tumor areas, etc [26], these problems all affect the accuracy of brain tumor segmentation.

In this article, to meet the above challenges, we use a combined model, integrate the "Squeeze and Excite" (SE) module and the "Attention Guide Filter" (AG) module into the VNet model for image segmentation of 3D MRI glioma brain tumors, it is an end-to-end network structure. We input data into the model in the form of volume input and use three-dimensional convolution to process MRI images. When the image is compressed along with different encoder blocks, the resolution is halved, and the number of channels increases. After the image is convolved, the compression and compression module is performed. The importance of each feature channel is automatically obtained through learning. Then according to this important level to promote useful functions, and cancel the less useful functions of the current task. Each decoder receives the characteristics of the corresponding stage of downsampling and decompresses the image, in the upsampling, the AG module is integrated, the Attention block is used to eliminate the influence of noise and irrelevant background, and the Guide Image Filtering is used to guide image features and structural information (edge information), it is worth mentioning that the idea of skip connection is used in the model to avoid the disappearance of the gradient. Besides, we also use the Categorical_Dice loss function as the optimization function of the model, which effectively solves the problem of pixel imbalance.

We tested the performance of this model on the Multimodal Brain Tumor

Segmentation Challenge (BraTS) 2020 dataset and compared it with the results of other teams participating in the challenge. The results show that our model has a good segmentation effect and has the potential for clinical trials. The innovations of this article are: 1) Clever use of channel relationships, using global information to enhance useful information in the channel, to suppress useless information in the channel. 2) The attention mechanism is added, and the network structure is also full of jump connections. The information extracted by the downsampling can be quickly captured to enhance the performance of the model. 3) Use the Categorical_Dice loss function to solve the problem of imbalance between foreground voxels and background voxels.

**3. Methodology**

*3.1. Method summary*

Our task is to segment multiple sequences of 3D MRI brain tumor images. In order to obtain good segmentation performance, we propose a new network structure called AGSE-VNet, which combines SE (Squeeze-and-Excitation) [27] module with AG (Attention Guided Filter) module [28] is integrated into the network structure, allowing the network to use global information to enhance useful feature channels selectively and suppress useless feature channels, cleverly solving the mutual dependence of feature maps, effectively suppressing the background information of the image, and enhancing the accuracy of model segmentation. In the next section, we will introduce the network structure of AGSE−VNet in detail.

*3.2. Squeeze-and-Excitation Blocks*

Fig. 2 is a schematic diagram of the SE module, which mainly includes the Squeeze module and the Excitation module. The core of the module is to recalibrate the characteristic response of the channel adaptively by explicitly modeling the interdependence between the channels. $F_{tr}$ in the figure is a standard convolution operation, as shown in formula (1), input as X, $X \in R^{Z' \times W' \times H' \times C'}$, where $Z$ is the depth, $H$ is the height $W$ is the width, $C$ is the number of channels, the output is $U$, $U \in R^{Z \times W \times H \times C}$, $v_c^s$ is a three-dimensional spatial convolution, $v_c$ means that each channel acts on the corresponding channel feature.

$$U_c = v_c \times X = \sum_{s=1}^{C'} v_c^s \times x^s \tag{1}$$

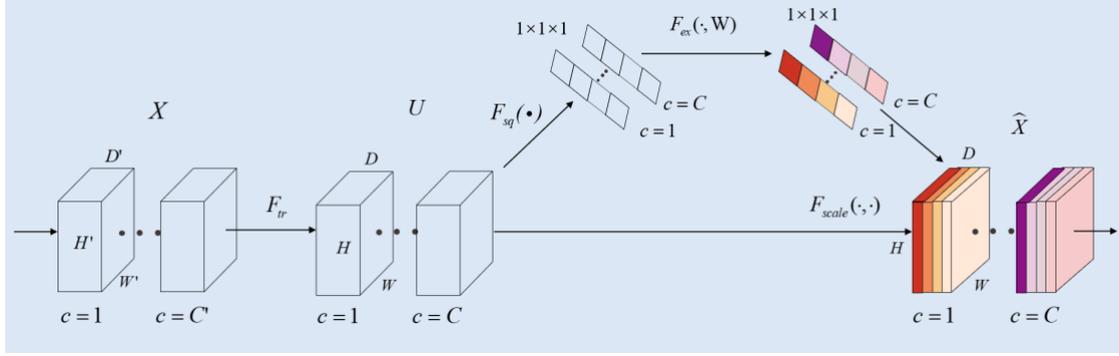

**Fig. 2.** SE network module diagram.

$F_{sq}(\cdot)$ is the squeeze operation. As shown in formula (2), the feature $U$ first passes the Squeeze operation. It compresses the features along the spatial gradient and aggregates the feature maps into the feature maps of dimension $W \times H$ as the feature descriptor. Each three-dimensional feature channel becomes a real number, which responds to the global distribution on the feature channel, to a certain extent, the real number at this time is closer to the global receptive field. This operation transforms the input of $H \times W \times C$ into $1 \times 1 \times C$ output.

$$z_c = F_{sq}(U_c) = \frac{1}{Z \times H \times W} = \sum_{k=1}^{Z} \sum_{i=1}^{W} \sum_{j=1}^{H} (k, i, j) \tag{2}$$

As shown in Equation 3, in order to limit the complexity and generalization of the model, using two fully connected layers as a parameterized gating mechanism. In Equation 3, $W_1 \times z$ represents a fully connected layer operation. The dimension of $W_1$ is $C \times \frac{C}{m}$. Here $m$ is a scaling parameter. In this article, we set $m = 4$ empirically. The parameter aims is to reduce the number of channels and thus reduce the number of calculations. Then through a ReLU layer, the output dimension remains unchanged and then multiplying with $W_2$. This process of multiplying with $W_2$ is also an operation of a fully connected layer. The dimension of $W_2$ is $C \times \frac{C}{m}$, and finally, through the Sigmoid function, the parameter $s$ is obtained.

$$s = F_{ex}(z, W) = \sigma(g(z, W)) = \sigma(W_2 \delta(W_1, z)) \qquad (3)$$

Where $\delta$ is the ReLU operation, $W_1 \in R^{\frac{C}{m} \times C}$, $W_2 \in R^{C \times \frac{C}{m}}$, and finally a $1 \times 1 \times C$ real number sequence is combined with $U$, recalibrated, and the final output is obtained by formula (4).

$$\widetilde{x_c} = F_{scale}(u_c, s_c) = s_c \cdot u_c \qquad (4)$$

Among them, $X = [x_1, x_2, ..., x_c]$ and $F_{scale}(u_c, s_c)$ refer to the corresponding channel between the feature map $u_c \in R^{W \times H}$ and the scalar $s_c$.

*3.3 Attention Guided Filter Blocks*

Attention Guided Filter (AG) module combines attention block and guided image filtering. The Attention Guided Filter filters the low-resolution feature maps and high-resolution feature maps to recover spatial information and merge structural information from feature maps of different resolutions. Fig. 3 is a schematic diagram of the Attention Block, where $O$ and $I$ are the input of the attention guided filter, and the attention map obtained by the calculation. Attention Block is extremely critical in this method. It effectively solves the influence of the background on the foreground and has the effect of highlighting the foreground and reducing the background. For the given feature maps $O$ and $I_l$, use convolution with a channel of $1 \times 1 \times 1$ to perform a linear transformation, and then combine the two converted feature maps with the ReLU layer through element addition, and then use a $1 \times 1 \times 1$. The convolution is again linearly transformed, and the sigmoid is most used to activate the final attention feature map $T$.

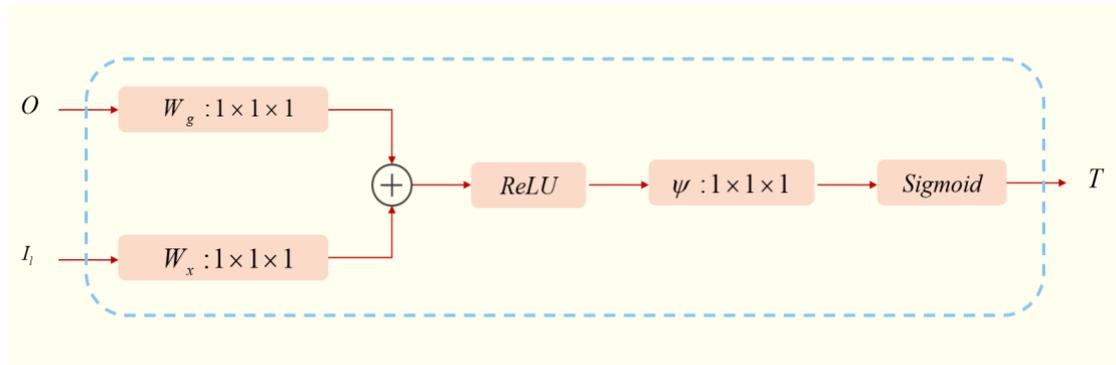

**Fig. 3.** Attention Block schematic diagram.

Fig. 4 is a schematic diagram of the results of the AG module. The input is the guided feature map $(I)$ and the filtered feature map $(O)$, and the output is the high-resolution feature map $(\tilde{O})$, which is the product of the joint action of $I$ and $O$, as shown in formula (5).

$$\tilde{O}_i = \sum_{i \in w_k} W_{ij}(I) \cdot O_j \tag{5}$$

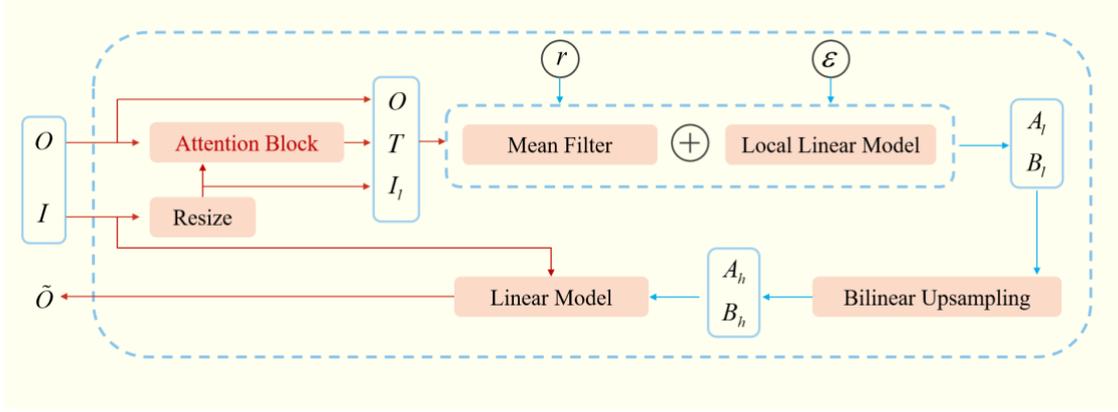

**Fig. 4.** Attention Guided Filter Blocks structure diagram.

Different from the guided filtering proposed by Kaiming He [29], the attention feature map $T$ is generated from the filtered feature map $(O)$ through the Attention Block module. First, the guided feature map $I$ is down-sampled to obtain a low-resolution feature map $I_l$, which is similar to the feature map $O$ Same size. Then minimize the reconstruction error of $O$ and $I_l$ to obtain the coefficients $A_l$ and $B_l$ of the attention guided filter. After that, by down-sampling $A_l$ and $B_l$ or coefficients $A_h$ and $B_h$, finally get the high resolution generated by the attention filter Feature map $\tilde{O}$. Among them, the attention filter is essentially a specific window $w_k$ with a radius of $r$. In particular, the attention guide filter will construct a square window $w_k$, and the radius of $k$ at each position is $r$. In our study, we set $r = 16$ and $\varepsilon = 0.1^2$ empirically based on the final segmentation performance. $(a_k, b_k)$ is also the only certain constant coefficient, as shown in formula (6), where ridge regression with a standard term is used to calculate the minimum reconstruction error.

$$\min_{a_k,b_k} E(a_k, b_k) := \sum_{i \in w_k}(T_i^2(a_k I_{li} + b_k - O_i)^2 + \varepsilon a_k^2) \tag{6}$$

Where $T_i$ is the attention weight at position $i$, $\varepsilon$ is the regularization parameter, and the calculation of $(a_k, b_k)$ is shown in formula (7).

$$a_k = \frac{\frac{1}{|w|}\sum_{i \in w_k}(I_i O_i - \mu_k \overline{O_k})}{\sigma_k^2 + \varepsilon}, b_k = \overline{O_k} - a_k \mu_k \tag{7}$$

Where $\mu_k$ is the average value of the window $w_k$ pixels in the image $I$, $\sigma_k^2$ is the variance of the window $w_k$, $|w|$ is the sum of the window pixels, and $O_k$ is the average pixel value $O_k = \frac{1}{|w|}\sum_{i \in w_k} O_i$ of the image $O$ to be filtered in the window $w_k$, so that the non-edge area can be found if a pixel is surrounded by multiple windows, calculate the average value of all windows containing the pixel at that pixel for such a pixel, as shown in formula (8).

$$O_i = \frac{1}{|w|}\sum_{k, i \in w_k}(a_k I_i + b_k) = A_l * I_l + B_l \tag{8}$$

Get $A_h$ and $B_h$ through upsampling, and finally get the output $\tilde{O} = A_h * I + B_h$.

*3.4. Downsamplings*

In Fig. 1, we provide a schematic diagram of AGSE-VNet. The network structure is divided into encoder and decoder in total, as shown in Fig. 5. Among them, Fig. a is the encoder, and the coding area mainly performs compression path, and Fig. b is the decoder, and the decoding area performs decompression. Downsampling is composed of four encoder blocks, each of which includes 2-3 layers of convolution, an extrusion and excitation layer and a downsampling layer, the processing process of the SE module is shown on the right side of Fig. 5(a). The feature extraction is performed by convolution with a step size of 2. The convolution is as follows (9) (10) shows:

$$i_s = i + (s - 1)(i - 1) \tag{9}$$

$$o = \left[\frac{i_s + 2p - k}{s} + 1\right] \tag{10}$$

Where $i$ is the input size, $i_s$ is the output size after filling, $s$ is the step size, $p$ is the filling size, $k$ is the convolution kernel size, and $o$ is the output size.

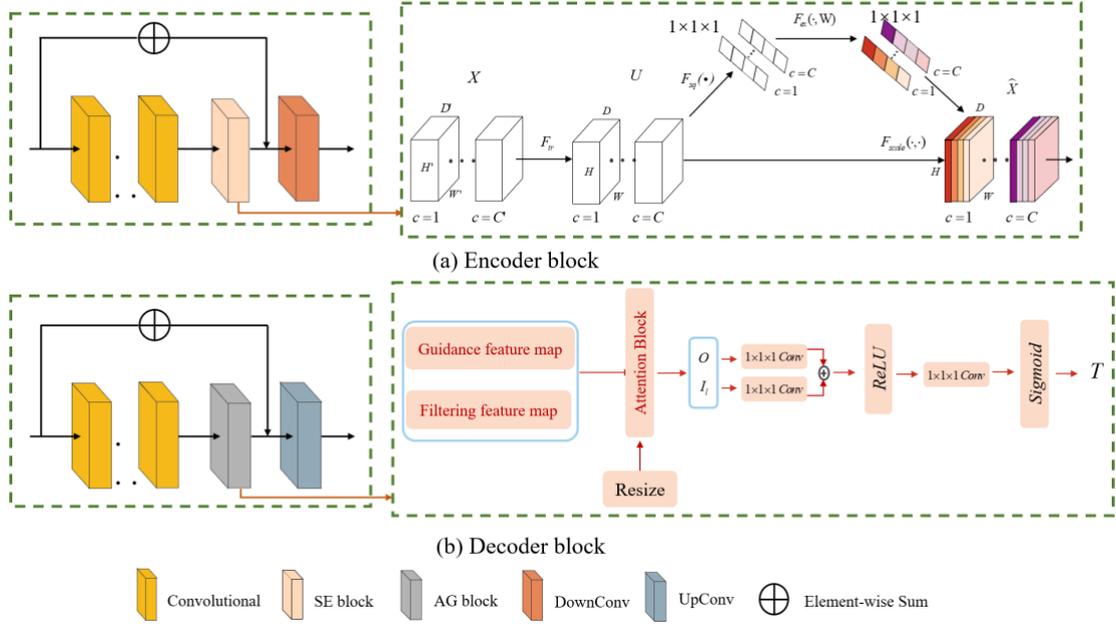

**Fig. 5.** The architecture of encoder block and decoder block with AGSE-VNet.

When the image is compressed along with different encoder blocks, its resolution is halved and the number of channels doubled. This is achieved by convolution of $3 \times 3 \times 3$ voxels with a step size of 2. After the convolution operation, the squeeze and compression module is performed, which ingeniously solves the relationship between the channels and improves the effective information transmission in the channels. It is worth mentioning that all convolutional layers have adopted normalization and dropout processing, and the ReLU activation function has also been applied to various positions in the network structure. Besides, a jump connection method is also used in the model to avoid the disappearance of the gradient as the network structure deepens.

*3.5. Upsampling*

After downsampling the model, we introduced the AG module to solve the problem of restoring spatial information and fusing structural information from low-resolution feature maps to high-resolution feature maps. The AG module is similar to the SE module. Based on not changing the dimensions of input and output, the features are enhanced. Therefore, we replace the splicing module in the VNet model with the AG module and integrate it into the decoder. The structure diagram is shown in Fig. 5.

Each decoder block includes an upsampling layer, an AG module, and three layers of convolution, the processing flow of the AG module is shown in the box on the right side of Fig.5(b). The decoder decompresses the image. In the up-sampling, this article uses deconvolution with a step size of 2 to fill in the image feature information. The deconvolution is shown in formula (11):

$$o = s(i - 1) + 2p - k + 2 \qquad (11)$$

Each decoder block receives the characteristics of the corresponding stage of downsampling. The convolution kernel used in the last layer of the network structure keeps the number of output channels consistent with the number of categories. Finally, the channel value is converted into a probability value output through the sigmoid function, and the voxel is converted into a brain tumor gangrene area. The idea of skip connection is adopted in each decoder block. The feature map after processing by the encoder and decoder is shown in Fig. 6., where Fig. 6(a) is a feature map processed by the encoder, and Fig. 6(b) is a feature map processed by the decoder.

*3.6. Skip connection*

To further make up for the information lost in the downsampling of the encoder, concat is used between the encoder and decoder of the network to fuse the feature maps of the corresponding positions in the two processes. In particular, the method extracted in this article uses the AG (Attention Guided Filter Blocks) module instead of concat, so that the decoder can obtain information during upsampling. With more high-resolution information, the detailed information in the original image can be restored more perfectly, and the segmentation accuracy can be improved.

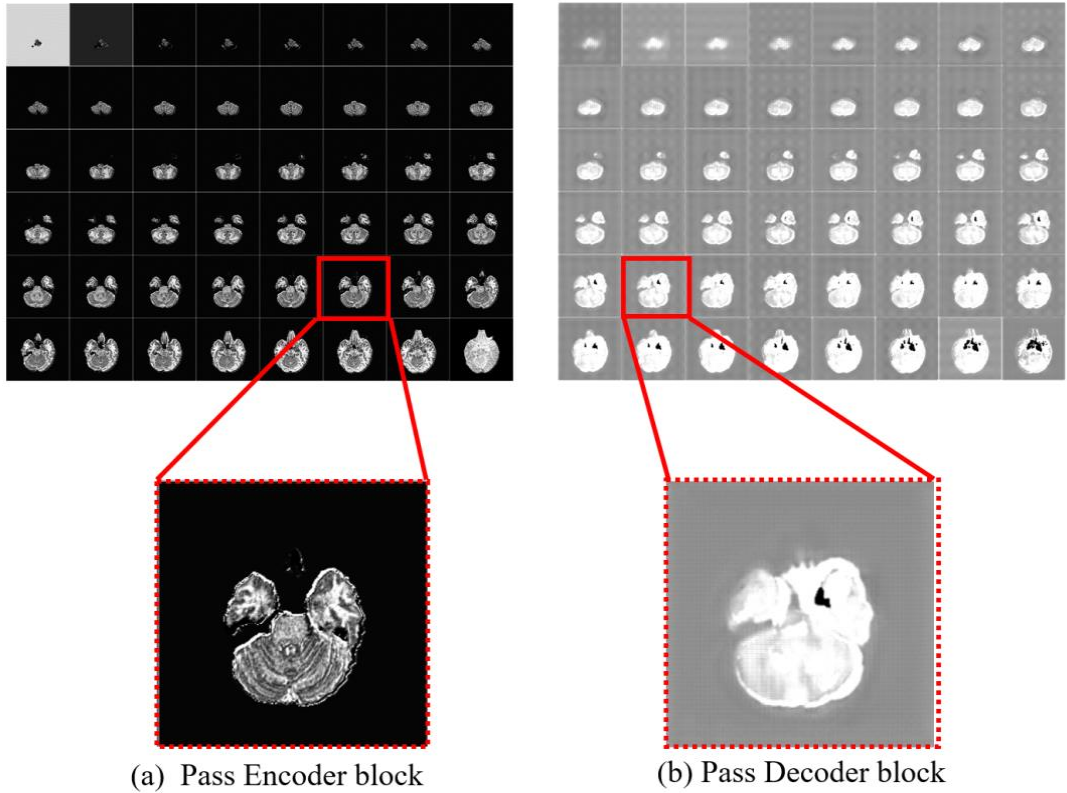

**Fig. 6.** Feature map processed by encoder and decoder.

We introduced adjacent layer feature reconstruction and cross-layer feature reconstruction in the network. The cross-layer feature reconstruction module is based on the encoder-decoder structure. In the process of network communication, as the network continues to deepen, the acceptance domain of the corresponding feature map will become larger and larger, but the retained detailed information will become less and less. Based on the encoder-decoder symmetric structure, the splicing layer is used to splice the feature maps extracted from the down-sampling in the encoder process and the new features obtained from the up-sampling in the decoder process to perform channel-dimensional splicing. Retaining more important feature information is conducive to achieving a better segmentation effect. Adjacent layer feature reconstruction is to establish a branch between each pair of adjacent convolutional layers with the same size feature map, that is, use the splicing layer to convolve the feature map obtained through the convolution of the previous layer and the next layer.

Obtaining the channel size achieves the purpose of maximizing the use of feature information in all previous layers.

*3.7. Loss function*

At present, the segmentation of medical images faces the problem of the imbalance between the foreground and the background regions. We also face such challenges in our tasks. Therefore, we choose the Categorical_Dice loss function as the optimization function of our model. Heavy to solve this problem by adjusting the weight of each forecast category. We set the weight of gangrene, edema, and enhanced tumor to 1, and the weight of background to 0.1. The Categorical_Dice loss function is shown in formula (12):

$$Dice(P,G) = \frac{2|P \cap G|}{|P|+|G|} \tag{12}$$

Among them, $G$ is Mask, the ground truth encoded by one-hot, $G \in [None, 64,128,128,4]$ and $P$ represents the predicted value, which is the probability result obtained after softmax calculation, $P \in [None, 64,128,128,4]$. The partial differential calculation of formula (13) is performed to obtain the gradient value relative to the predicted $j$-th voxel, where N stands for voxel, $p_i \in P$ and $g_i \in G$.

$$\frac{\partial D}{\partial p_j} = 2\left[\frac{g_j(\sum_i^N p_i^2 + \sum_i^N g_i^2) - 2p_j(\sum_i^N p_i g_i)}{(\sum_i^N p_i^2 + \sum_i^N g_i^2)^2}\right] \tag{13}$$

The weight distribution of the loss function of each node is shown in formula (14), and the weight value is $[0.1,1.0,1.0,1.0]$.

$$Loss = -Dice(P,G) \times weight \tag{14}$$

## 4. Materials

*4.1. Dataset*

In this research, we use the dataset of the BraTS 2020 challenge to train and test our model [30-31]. The data set contains two types, namely low-grade glioma (LGG) and glioblastoma (HGG), each category has four modal images: T1 weighting (T1), T1-weighted contrast-enhanced (T1-CE), and T2 weighting (T2), fluid attenuation inversion recovery (FLAIR). The mask of the brain tumor includes the gangrene area, edema area, and enhancement area. Our task is to segment the three sub-regions formed by nesting tags, which are enhancement tumor (ET), whole tumor (WT), and tumor core (TC).

There are 369 cases in the training set and 125 cases in the validation set. The masks corresponding to these cases are not used for training, and their functions are mainly used for evaluating the model after training.

*4.2. Design Detail*

In deep learning training, the setting of hyperparameters is very essential, and it will determine the performance of our model. But often in training, the initial value of the hyperparameter is set by experience. In the training of the AGSE-VNet model, the initial learning rate is set to 0.0001, the dropout is set to 0.5, the number of training steps is about 350,000, and then the learning rate is adjusted to 0.00003. The dataset is halved every time it is traversed, and the data is shuffled to enhance the robustness and generalization ability of the model.

The experimental environment was conducted on Tensorflow 1.13.1, and the runtime platform processor was Intel (R) Core (TM) i7-9750H CPU @ 2.60GHz, 32GB RAM, Nvidia GeForce RTX 2080, 64-bit Windows 10. The development software platform is PyCharm, and the python version is 3.6.9.

*4.3. Pre-Processing*

Since our data set has four modalities, T1, T1-CE, T2, and FLAIR, there is a problem of different contrast, which may cause the gradient to disappear during the training process, so we use standardization to process the image, from the image pixel. The image data is normalized by subtracting the average value and dividing by the standard deviation. Calculated as follows:

$$\hat{X} = \frac{X-\mu}{\sigma} \quad (15)$$

where, $\mu$ donates the mean of the image, $\sigma$ donates standard deviation, $X$ donates the image matrix, $\hat{X}$ is the normalized image matrix.

After normalization, we merge the images of the four modalities with the same contrast to form a three-dimensional image with four channels. The original image size is, and the combined image size becomes. The size of the label is, and its pixel value contains 4 different values. Channel 0 is the normal tissue area, 1 is the gangrene area, 2 is the edema area, and 3 is the enhanced tumor area. Then, divide the image and mask into multiple blocks and perform the patch operation. Each case generates 175 images with a size of 128×128×64. Finally, save it in the corresponding folder in NumPy .npy format (https://numpy.org/doc/stable/reference/). The preprocessed image is shown in Fig. 7.

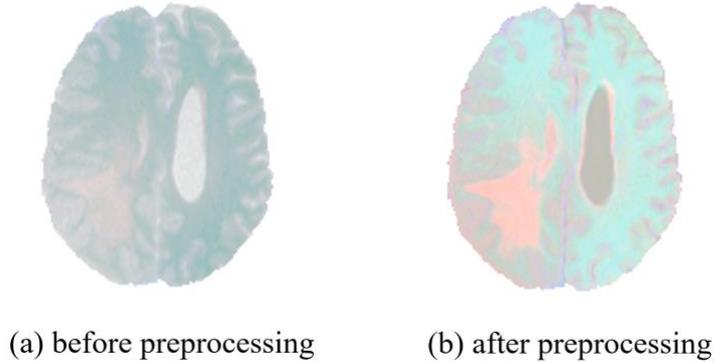

(a) before preprocessing     (b) after preprocessing

**Fig. 7.** Preprocessed result.

*4.4. Evaluation Metrics*

We use the dice coefficient, specificity, sensitivity, and Hausdorff95 distance to measure the performance of our model. Dice coefficient is calculated as:

$$Dice = \frac{2TP}{FN+FP+2TP} \quad (16)$$

where $TP$, $FP$ and $FN$ are the number of true positive, false positive, and false negative respectively. Specificity can be used to evaluate the number of true negative and false positive, it is used to measure the model ability to predict the background area, defined as:

$$Specificity = \frac{TN}{TN+FP} \tag{17}$$

where $TN$ is the number of true negative. Sensitivity can be used to evaluate the number of the true positive and false negative, it is used to measure the sensitivity of the model to segmented regions, defined as:

$$Sensitivity = \frac{TP}{TP+FN} \tag{18}$$

The Hausdorff95 distance measures the distance between the surface of the real area and the predicted area, which is more sensitive to the segmented boundary, defined as:

$$Haus95(T,P) = \max\left\{\sup_{t\in T, p\in P} \inf d(t,p), \sup_{p\in P, t\in T} \inf d(t,p)\right\} \tag{19}$$

where inf denotes the infimum and sup denotes the supremum, $t$ and $p$ donate the points on the surface $T$ of the ground-truth area and the surface $P$ of the pre dataset dictated area. Besides, $d(\cdot,\cdot)$ calculates the distance between the assembly point $t$ and the assembly point $p$.

## 5. Results and Discussions

*5.1. Results on AGSE-VNet model.*

Our data set includes a training set and a test set. The training set contains 369 cases and the test set contains 125 cases. The mask of the tumor includes the gangrene area, edema area, enhancement area, and background area. The labels correspond to 1, 2, 4, and 0, respectively. These labels are merged into three nested sub-areas, namely the enhancing tumor (ET), the whole tumor (WT), and the tumor core (TC), for these sub-regions, we use four indicators of sensitivity, specificity, dice coefficient, and Hausdorff95 distance to measure the performance of the model. We use the data set of BraTS 2020 for training and verification, and the average index obtained is shown in

Table 1. From Table 1, we observe that the model has a better segmentation effect on the WT region. The Dice and Sensitivity of the training set and the validation set are 0.846, 0.849, 0.825, and 0.833, respectively, which are significantly better than other regions.

**Table 1:** Quantitative valuation on the training set and validation set.

|  | Dice | | | Sensitivity | | | Specificity | | | Hausdorff95 | | |
| --- | --- | --- | --- | --- | --- | --- | --- | --- | --- | --- | --- | --- |
|  | ET | WT | TC | ET | WT | TC | ET | WT | TC | ET | WT | TC |
| **Training** | 0.70 | 0.85 | 0.77 | 0.72 | 0.83 | 0.74 | 0.99 | 0.99 | 0.99 | 35.70 | 8.96 | 17.40 |
| **Validation** | 0.68 | 0.85 | 0.69 | 0.68 | 0.83 | 0.65 | 0.99 | 0.99 | 0.99 | 47.40 | 8.44 | 31.60 |

On this basis, we conduct a statistical analysis of the experimental results. Fig. 8 and Fig. 9 are the scatter plots and box plots of the four evaluation indicators of the training set and test set, reflecting the distribution characteristics of the results. It can be seen from the box plot that there are fewer outliers of various indicators and minimal fluctuation of results. The horizontal line in the box plot represents the median of this set of data. It can be observed that the three indicators of Dice, Sensitivity, and Specificity are at a higher level, which shows that the segmentation effect of our proposed model is located in a higher area. In the results of the four indicators, the sensitivity results are all concentrated at a higher level. It can be seen that the fluctuation range is small. Observing the scatter diagram on the left side, it can be seen that the data are all clustered at a higher position, indicating that our model is the background area has a high level of prediction, which can effectively alleviate the problem of imbalance between foreground pixels and background.

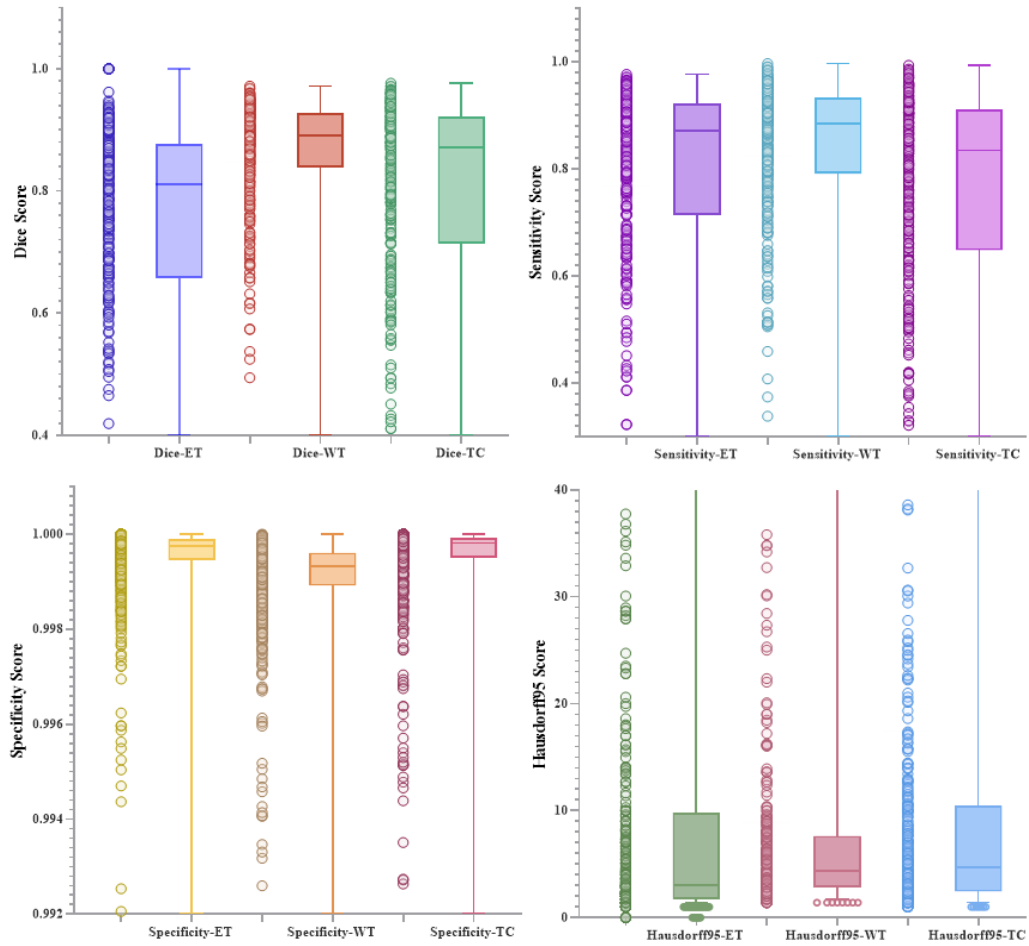

**Fig. 8.** A collection of scatter plots and box plots of four indicators in the training set.

We randomly selected several slices from the training set and compared the actual situation with the results predicted by our model, as shown in Fig. 10(a), the first line is the original image, the second line is the label, and the third line is the tumor sub-region predicted by our model. At the same time, we also selected two of them to display in Fig. 10(b). Among them, the green area is the whole tumor (WT), the red area is the tumor core (TC), and the area combining yellow and red represents the enhancing tumor(ET). We show the 3D image of the segmentation result in the last two columns. From the comparison of the segmentation results, we can find that our model has a good effect on brain tumor segmentation, especially the whole tumor (WT) region segmentation effect is excellent. However, the segmentation prediction of the tumor core (TC) is slightly biased, which may not be suitable for extraction due to the small features of the tumor core.

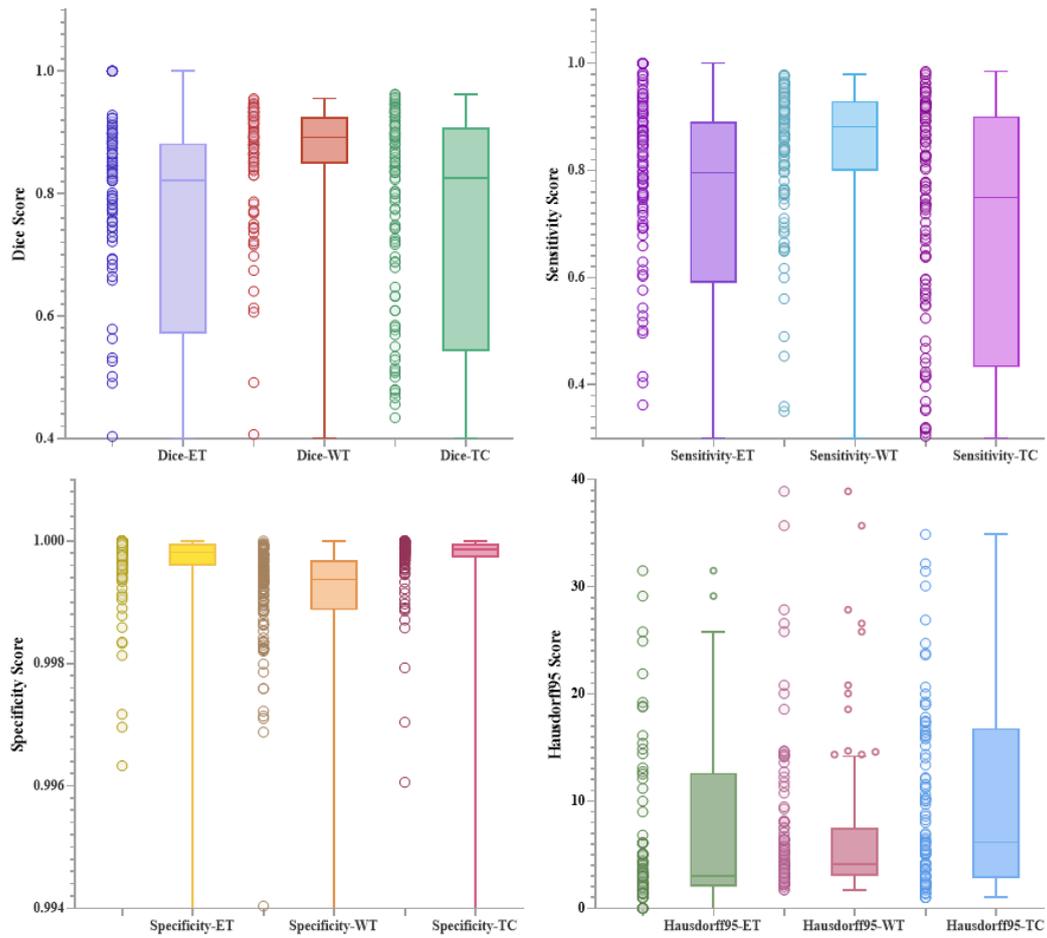

**Fig. 9.** A collection of scatter plots and box plots of four indicators in the validation set.

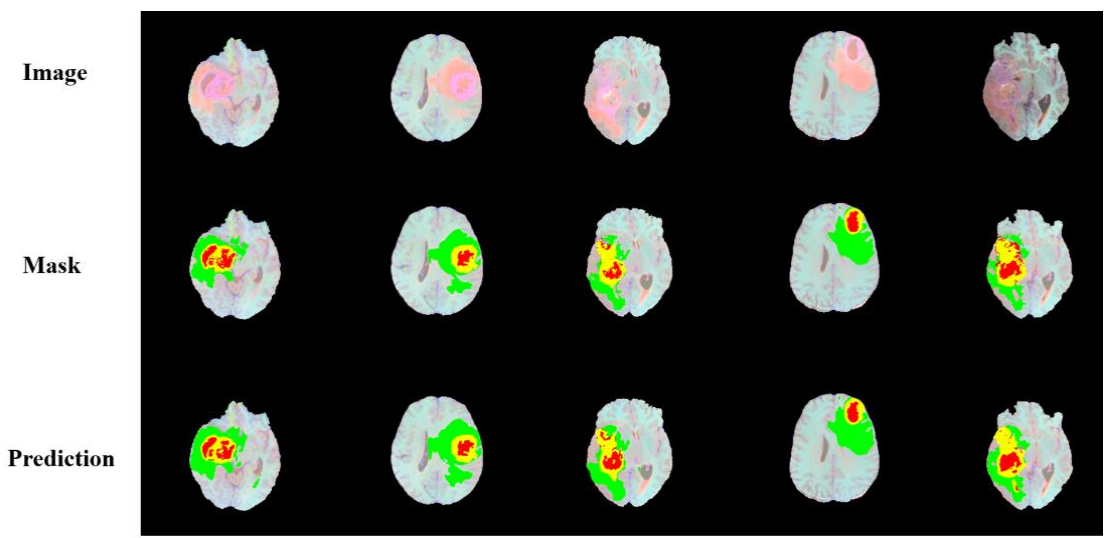

(a)

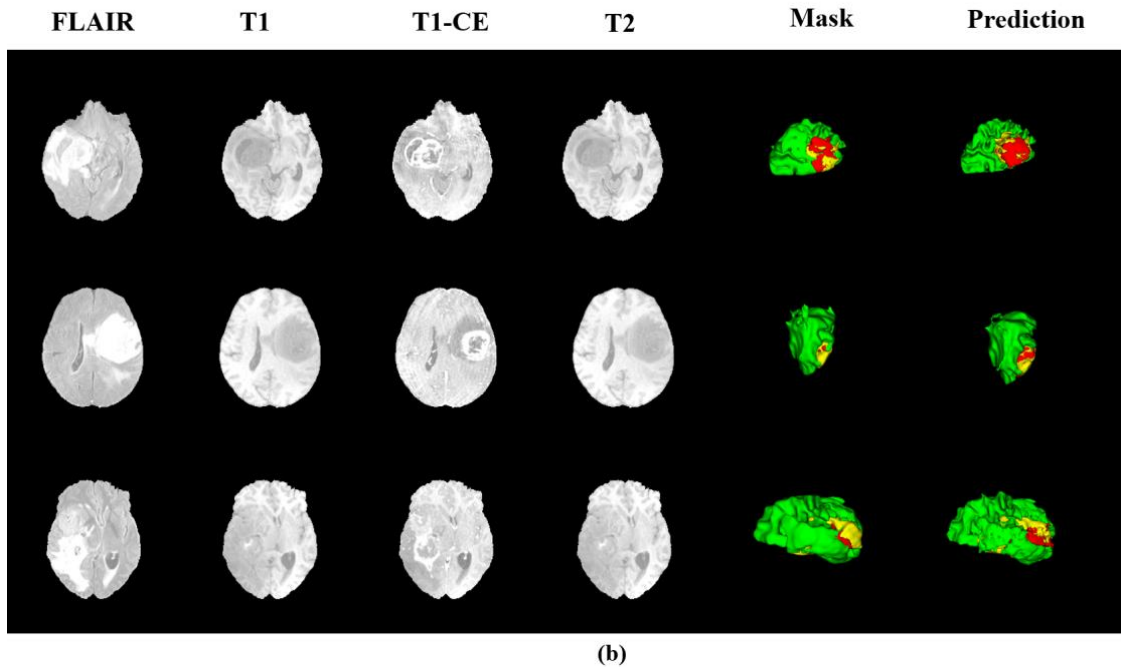

(b)

**Fig. 10.** Display of segmentation results in the training set.

After the training is completed, we randomly selected several segmentation slices in the validation set for display, as shown in Fig. 11(a). Similarly, in Fig. 11(b), we also show the three-dimensional image of the segmentation result and annotate the accuracy value of the ET region, as can be seen from the figure, our model has a good segmentation effect for MRI images of different intensities, and can accurately segment tumor sub-regions, which has a certain potential in brain tumor image segmentation.

In our research, we proposed the AGSE-VNet model to segment 3D MRI brain tumor images and obtained better segmentation results on the BraTS 2020 dataset. In order to further verify the effect of our segmentation, compare our experimental method with the methods proposed by other outstanding teams participating in the competition. The results of other teams are available on the official website of the BraTS Multimodal Brain Tumor Segmentation Challenge 2020 (https://www.cbica.upenn.edu/BraTS20/lboardTraining.html). The comparison results of the training set are shown in Table 2, and the comparison results of the verification set are shown in Table 3. From the results in the table, we can find that our model performs well in the whole tumor (WT) region and obtains relatively excellent results, indicating that the method we proposed has a certain potential in segmentation.

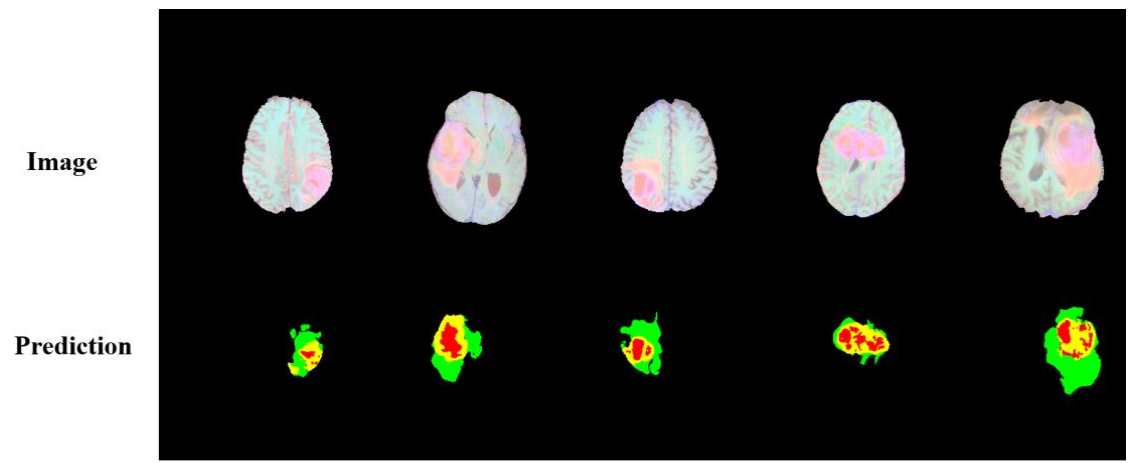

(a)

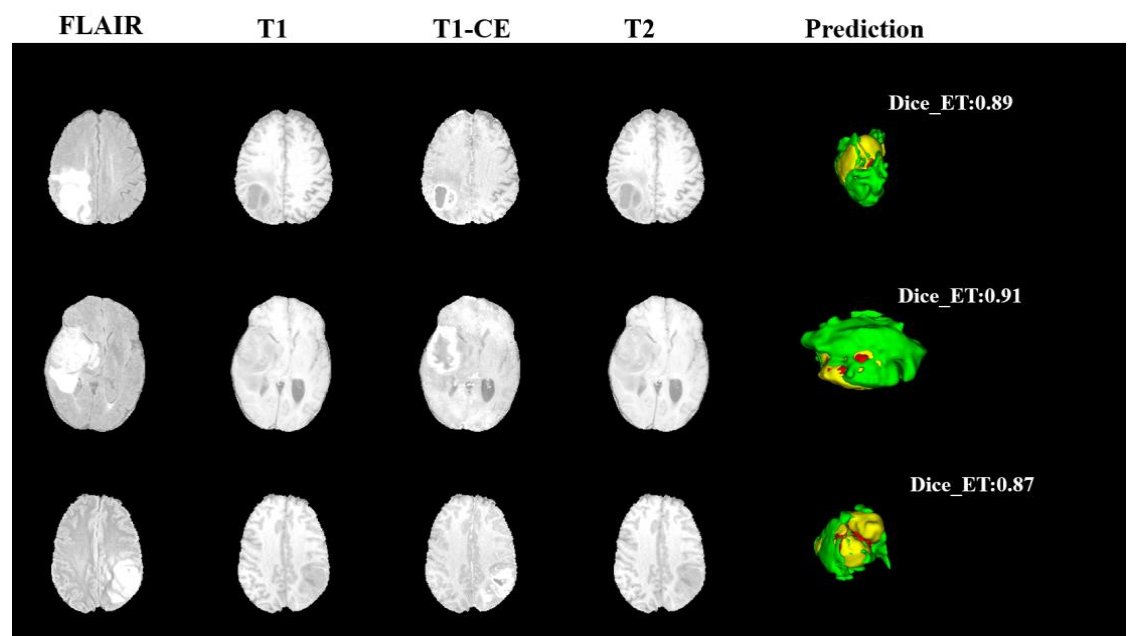

(b)

**Fig. 11.** Display of segmentation results in the validation set.

**Table 2:** The results of various indicators in the training set

| Team | Dice | | | Sensitivity | | | Specificity | | | Hausdorff95 | | |
|---|---|---|---|---|---|---|---|---|---|---|---|---|
| | ET | WT | TC | ET | WT | TC | ET | WT | TC | ET | WT | TC |
| proposed | 0.70 | 0.85 | 0.77 | 0.72 | 0.83 | 0.74 | 0.99 | 0.99 | 0.99 | 35.70 | 8.96 | 17.40 |
| mpstanford | 0.60 | 0.78 | 0.72 | 0.56 | 0.80 | 0.75 | 0.99 | 0.99 | 0.99 | 35.95 | 17.68 | 17.21 |
| agussa | 0.67 | 0.87 | 0.79 | 0.69 | 0.87 | 0.82 | 0.99 | 0.99 | 0.99 | 39.25 | 15.75 | 17.05 |
| ovgu_seg | 0.65 | 0.81 | 0.75 | 0.72 | 0.78 | 0.76 | 0.99 | 0.99 | 0.99 | 34.79 | 9.50 | 8.93 |
| AI-Strollers | 0.59 | 0.73 | 0.61 | 0.52 | 0.73 | 0.64 | 0.99 | 0.97 | 0.98 | 38.87 | 20.81 | 24.22 |
| uran | 0.48 | 0.79 | 0.64 | 0.45 | 0.74 | 0.61 | 0.99 | 0.99 | 0.99 | 37.92 | 7.72 | 14.07 |
| CBICA | 0.54 | 0.78 | 0.57 | 0.64 | 0.82 | 0.53 | 0.99 | 0.99 | 0.99 | 20.00 | 46.30 | 39.60 |
| unet3d-sz | 0.69 | 0.81 | 0.75 | 0.77 | 0.93 | 0.83 | 0.99 | 0.96 | 0.98 | 37.71 | 19.57 | 18.36 |
| iris | 0.76 | 0.88 | 0.81 | 0.78 | 0.90 | 0.83 | 0.99 | 0.99 | 0.99 | 32.30 | 18.07 | 14.70 |
| VuongHN | 0.74 | 0.81 | 0.82 | 0.84 | 0.98 | 0.84 | 0.95 | 0.93 | 0.99 | 21.97 | 12.32 | 8.72 |

**Table 3:** The results of various indicators in the validation set

| Team | Dice | | | Sensitivity | | | Specificity | | | Hausdorff95 | | |
|---|---|---|---|---|---|---|---|---|---|---|---|---|
| | ET | WT | TC | ET | WT | TC | ET | WT | TC | ET | WT | TC |
| proposed | 0.68 | 0.85 | 0.69 | 0.68 | 0.83 | 0.65 | 0.99 | 0.99 | 0.99 | 47.40 | 8.44 | 31.60 |
| mpstanford | 0.49 | 0.72 | 0.62 | 0.49 | 0.81 | 0.69 | 0.99 | 0.99 | 0.99 | 61.89 | 26.00 | 28.02 |
| agussa | 0.59 | 0.83 | 0.69 | 0.60 | 0.87 | 0.71 | 0.99 | 0.99 | .0.99 | 56.58 | 23.23 | 29.59 |
| ovgu_seg | 0.60 | 0.79 | 0.68 | 0.66 | 0.79 | 0.67 | 0.99 | 0.99 | 0.99 | 54.07 | 12.05 | 19.10 |
| AI-Strollers | 0.58 | 0.74 | 0.61 | 0.52 | 0.77 | 0.62 | 0.99 | 0.99 | 0.99 | 47.23 | 24.03 | 31.54 |
| uran | 0.75 | 0.88 | 0.76 | 0.77 | 0.85 | 0.71 | 0.99 | 0.99 | 0.99 | 36.42 | 6.62 | 19.30 |
| CBICA | 0.63 | 0.82 | 0.67 | 0.76 | 0.78 | 0.75 | 0.99 | 0.99 | 0.99 | 9.60 | 10.70 | 28.20 |
| unet3d-sz | 0.70 | 0.84 | 0.72 | 0.71 | 0.87 | 0.79 | 0.99 | 0.99 | 0.99 | 42.09 | 10.48 | 12.32 |
| iris | 0.68 | 0.86 | 0.73 | 0.67 | 0.90 | 0.70 | 0.99 | 0.99 | 0.99 | 44.13 | 23.87 | 20.02 |
| VuongHN | 0.79 | 0.90 | 0.83 | 0.80 | 0.89 | 0.80 | 0.99 | 0.99 | 0.99 | 21.43 | 6.74 | 7.05 |

*5.2. Discussion*

The method proposed in this paper cleverly solves the problem of interdependence between channels, and autonomously extracts effective features from channels to suppress useless feature channels. After the features extracted by the encoder, low-resolution feature maps and high-resolution feature maps are filtered through the Attention module, to recover spatial information and fusion structural information from feature maps of different resolutions, our method is not affected by the size and location of the tumor. For MRI images of different intensities, the tumor area can be automatically identified, and the tumor sub-regions can be feature extracted and segmented, and the segmentation effect obtained has a good performance. This is beneficial to radiologists and oncologists, who can quickly predict the condition of the tumor and assist in the treatment of the patient. Comparing the results in Table 2 and Table 3, we find that our model performs well in the whole tumor (WT) area, but does not perform well in the enhancing tumor (ET) and the tumor core (TC) areas, this may be because the target in the ET area is small and the feature is fuzzy and difficult to extract. At the same time, we compare our method with some classic algorithms for brain tumor segmentation. The results are shown in Table 4. In the BraTS Challenge, 2018, zhou et al.[32] and others proposed a lightweight one-step multi-task segmentation model, by learning the shared parameters of joint features and the composition features of distinguishing specific task parameters, the imbalance factors of tumor types are effectively alleviated, uncertain information is suppressed, and the segmentation result is improved. In the method proposed by Zhao et al., a new segmentation framework was developed, using a fully convolutional neural network to assign different labels to the image in pixel units, optimize the output results of FCNNs by using the recurrent neural network constructed by the conditional random place, this method was verified on the BraTS 2016 dataset and got a good segmentation effect. Pereira et al. proposed an automatic positioning method for convolutional neural networks, which achieved good results in the BraTS 2015 dataset.

**Table 4:** Comparison of our proposed AGSE-VNet model with classic methods

| Method | Dice_ET | Dice_WT | Dice_TC | Dataset |
|---|---|---|---|---|
| **Proposed** | 0.67 | 0.85 | 0.69 | BraTs 2020 |
| **Zhou et al.** | 0.65 | 0.87 | 0.75 | BraTs 2018 |
| **Zhao et al.** | 0.62 | 0.84 | 0.73 | BraTs 2016 |
| **Pereira et al.** | 0.65 | 0.78 | 0.75 | BraTs 2015 |

Analyzing Table 4, we found that our model has certain advantages in segmentation, there are still differences in TC regional accuracy, and the model has limitations. In future work, we will propose solutions to this situation, such as how to further segment the region of interest after our model has extracted it, in order to improve the accuracy of the enhancing tumor (ET) and the tumor core (TC) areas, more characteristic information can be captured. Besides, the algorithms proposed in many top methods have their areas of excellent performance. How we combine the advantages of these algorithms and integrate them into our model is the focus of our future work. In clinical treatment, it helps experts to understand the patient's current situation more quickly and accurately, saving experts time, and realizing a leap in the quality of automatic medical segmentation.

In addition, in order to verify the robustness of our model to resist noise interference, we have now added Gaussian noises in the frequency domain (k-space) of the testing data to simulate realistic noise contaminations. The comparison results are shown in Fig.12. From the noisy and no-noise segmentation results, we have found that the segmentation results of our AGSE-VNet model for the three regions are not much different. These results can demonstrate that our model has a significant advantage in generalization when noises are present.

## 6. Conclusion

All in all, we have implemented a good method to segment 3D MRI brain tumor images, this method can automatically segment the three regions of the enhancing tumor (ET), the whole tumor (WT), and the tumor core (TC) of the brain tumors. We conducted experiments on the BraTS 2020 data set and got good results. The AGSE-VNet model is improved based on VNet. There are five encoder blocks and four decoder blocks. Each encoder block has an extrusion and excitation block, and each decoder has an Attention Guild Filter block. Such a design can be embedded in our model without affecting the size mismatch of the network structure under the condition that the input ratio and output ratio are unchanged. After the SE module processes the model, the network learns the global information and selects the useful information in the enhancement channel, and then uses the attention mechanism of the Attention Guild Filter block to quickly capture its dependencies and enhance the performance of the model. Secondly, we also introduced a new loss function Categorical_Dice, set different weights for unused masks, set the weight of the background area to 0.1, and set the tumor area of interest to 1, Ingeniously solve the problem of the voxel imbalance between the foreground and the background. Our online verification tool on the BraTS Challenge website evaluated this approach. It is found that our model is still different from the top methods for the segmentation of the enhancing tumor (ET) and the tumor core (TC) regions. It may be because the features of these two regions are small and difficult to extract. How to improve the accuracy of these two regions is our future work direction.

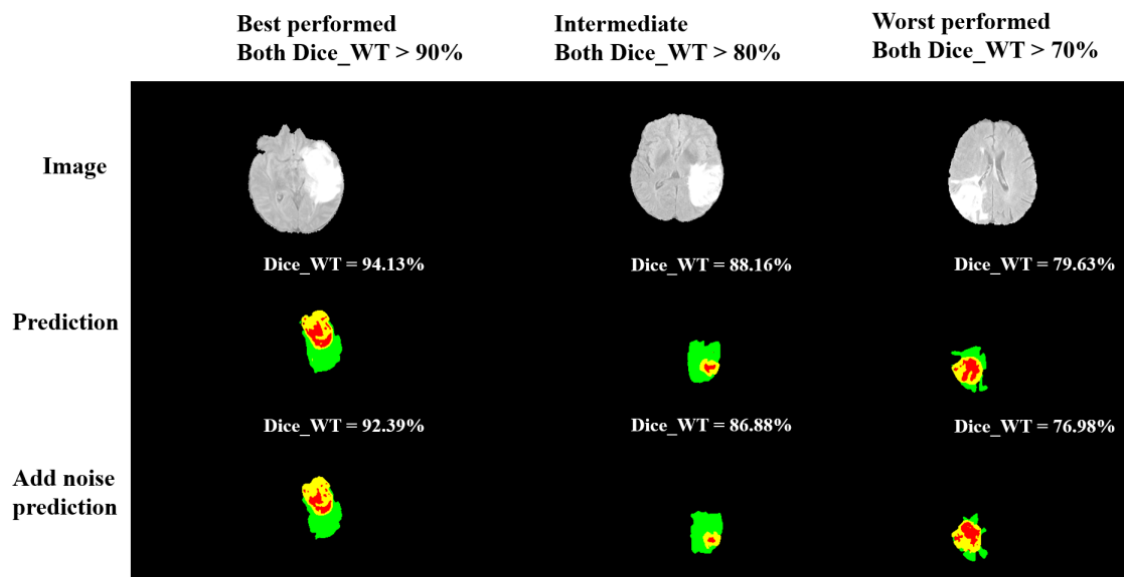

**Fig. 12.** Comparison of segmentation results without noise and noise added.

The automatic segmentation of brain tumors in the medical field has been a long-term research problem. How to design an automatic segmentation algorithm with short time and high accuracy, and then form a complete system is the current direction of a large number of researchers. Therefore, we must continue to optimize our segmentation model to achieve a qualitative leap in the field of automatic segmentation.

**Ethics approval and consent to participate**

Not applicable.

**Consent for publication**

- All authors contributed to the article and approved the submitted version.

**Availability of data and materials**

The datasets analysed during this current study are available in the BRATS 2020.

https://www.med.upenn.edu/cbica/brats2020/data.html

**Conflict of interest**

Not applicable.

**Funding**

This work is funded in part by the National Natural Science Foundation of China (Grants No. 62072413, 61602419), in part by the Natural Science Foundation of Zhejiang Province of China (Grant No. LY16F010008), in part by Medical and Health Science and Technology Plan of Zhejiang Province of China (Grant No. 2019RC224), in part by the Teacher Professional Development Project of Domestic Visiting Scholar in Colleges and Universities of Zhejiang Province of China (Grants No.2020-19, 2020-20), and also supported in part by the AI for Health Imaging Award 'CHAIMELEON: Accelerating the Lab to Market Transition of AI Tools for Cancer Management' [H2020-SC1-FA-DTS-2019-1 952172].

**Authors' contributions**

- XG, GY, and XL conceived and designed the study, contributed to data analysis, contributed to data interpretation, and contributed to the writing of the report. XG, GY, JY, WY, XX, WJ, and XL contributed to the literature search. JY and WY contributed to data collection. XG, GY, XX, WJ, and XL performed data curation and contributed to the tables and figures. All authors contributed to the article and approved the submitted version.

**References**

[1] Nie J, Xue Z, Liu T, Young GS, Setayesh K, Lei. G. Automated Brain Tumor Segmentation Using Spatial Accuracy-Weighted Hidden Markov Random Field. Comput Med Imaging Graph. 2009; 33(6): 431–441.

[2] Bakas S, Reyes M, Jakab A, et al. Identifying the Best Machine Learning Algorithms for Brain Tumor Segmentation. Progression Assessment, and Overall Survival Prediction in the BRATS Challenge, arXiv preprint arXiv:1811.02629. 2018.

[3] Essadike A, Ouabida E, Bouzid. A. Brain tumor segmentation with Vander Lugt correlator based active contour. Computer Methods and Programs in Biomedicine. 2018; 60:103–117.

[4] Havaei M, Davy A, Warde-Farley D, Biard A, Courville A, Bengio Y. Brain Tumor Segmentation with Deep Neural Networks. Medical Image Analysis. 2017; 35:18-31.

[5] Akkus Z, Galimzianova A, Hoogi A, Daniel R. Deep Learning for Brain MRI Segmentation: State of the Art and Future Directions. Journal of Digital Imaging. 2017; 30(4): 449-459.

[6] Hussain S, Anwar S, Majid M. Segmentation of Glioma Tumors in Brain Using Deep Convolutional Neural Network. Neurocomputing. 2017; 282.

[7] Sauwen N, Acou M, Cauter S, Sima DM, Veraart J, Maes F, Himmelreich U, Achten E, Van Huffel S. Comparison of unsupervised classification methods for brain tumor segmentation using multi-parametric MRI. NeuroImage: Clinical. 2016; 12(2):753-764.

[8] Milletari F, Navab N, Ahmadi S. V-Net: Fully Convolutional Neural Networks for Volumetric Medical Image Segmentatio. 2016 Fourth International Conference on 3D Vision (3DV). IEEE. 2016.

[9] Rickmann A, Roy A, Sarasua I, Navab N, Wachinger C. `Project & Excite' Modules for Segmentation of Volumetric Medical Scans. Image and Video Processing. 2019.

[10] Tustison N, Shrinidhi K, Wintermark M, Durst CR, Kandel BM, Gee JC, Grossman MC, Avants BB. Optimal Symmetric Multimodal Templates and Concatenated Random Forests for Supervised Brain Tumor Segmentation (Simplified) with ANTsR. Neuroinformatics. 2015; 13(2):209-225.

[11] Rose S, Crozier S, Bourgeat P, Dowson N，Salvado O，Raniga P，Pannek K，Coulthard A，Fay M，Thomas P. Improved delineation of brain tumour margins using whole-brain track-density mapping. Ismrm-esmrmb Joint Meeting: Clinical Needs & Technological Solutions. International Society of Magnetic Resonance in Medicine. 2009.

[12] Amiri S, Mahjoub MA, Rekik I. Bayesian Network and Structured Random Forest Cooperative Deep Learning for Automatic Multi-label Brain Tumor Segmentation. 10th International Conference on Agents and Artificial Intelligence. 2018.

[13] Balafar M. Fuzzy cc-mean based brain MRI segementation algorithms. Artif Intell Rev. 2014; 41(3):441-449.

[14] Pereira S，Pinto A，Alves V. Brain Tumor Segmentation Using Convolutional Neural Networks in MRI Images. IEEE Transaction on medical imaging. 2016; 35(5):1240-1251.

[15] Hao D, Yang G, Liu F, Mo Y, Guo Y. Automatic brain tumor detection and segmentation using U-Net based fully convolutional networks. Annual conference on medical image understanding and analysis. Springer, Cham. 2017.

[16] Wang G, Li W, Ourselin S, Vercauteren T. Automatic Brain Tumor Segmentation using Cascaded Anisotropic Convolutional Neural Networks. Computer Vision and Pattern Recognition. 2017;12 (5).

[17] Myronenko A. 3D MRI brain tumor segmentation using autoencoder regularizatio. Springer. 2018.

[18] Xue F, Nicholas T, Meyer C. Brain Tumor Segmentation using an Ensemble of 3D U-Nets and Overall Survival Prediction using Radiomic Features. Computer Vision and Pattern Recognition. 2018;279-288.

[19] NabilIbtehaz M, Rahman S. MultiResUNet: Rethinking the U-Net Architecture for Multimodal Biomedical Image Segmentation. Computer Vision and Pattern

Recognition. 2019;121.

[20] Xu C, Xu L, Ohorodnyk P, Roth M, Li M. Contrast agent-free synthesis and segmentation of ischemic heart disease images using progressive sequential causal GANs. Medical Image Analysis 101668. 2020.

[21] Zhou X, Li X, Hu K, Zhang Y, Chen Z, Gao X. ERV-Net: An efficient 3D residual neural network for brain tumor segmentation. Expert Systems with Applications. 2021;170.

[22] Saman S, Narayanan S. Active contour model driven by optimized energy functionals for MR brain tumor segmentation with intensity inhomogeneity correction. Multimedia Tools and Applications. 2021;80(4):21925-21954.

[23] Liu H, Li Q, Wang L. A Deep-Learning Model with Learnable Group Convolution and Deep Supervision for Brain Tumor Segmentation. Mathematical Problems in engineering. 2021;(3):1-11.

[24] Yurttakal A, Erbay H. Segmentation of Larynx Histopathology Images via Convolutional Neural Networks. Intelligent and Fuzzy Techniques: Smart and Innovative Solutions. 2021;949-954.

[25] Zhao X, Wu Y, Song G, Li Z, Zhang Y, Fan Y. A deep learning model integrating fcnns and crfs for brain tumor segmentation. Medical Image Analysis. 2018;43:98-111.

[26] Sturm D, Pfister S, Dtw J. Pediatric Gliomas: Current Concepts on Diagnosis, Biology, and Clinical Management. Journal of Clinical Oncology. 2017;35(21):2370.

[27] Hu J, Li S, Albanie S, Sun G. Squeeze-and-Excitation Networks. IEEE transactions on pattern analysis and machine intelligence. 2017;99.

[28] Zhang S, Fu H, Yan Y, Zhang Y, Wu Q, Tan M, Xu Y. Attention Guided Network for Retinal Image Segmentation. Medical Image Computing and Computer Assisted Intervention – MICCAI 2019. 2019.

[29] He K, Sun J, Tang X. Guided Image Filtering. Lecture Notes in Computer Science. 2013;35(6):1397-1409.

[30] Menze B, Jakab A, Bauer S, Jayashree KC, Keyvan F, Justin K. The Multimodal

Brain Tumor Image Segmentation Benchmark (BRATS). IEEE Transactions on Medical lmaging. 2015; 34(10):1993-2024.

[31] Bakas S, Akbari H, Sotiras A, Bilello M, Rozycki M, Kirby JS. Advancing The Cancer Genome Atlas glioma MRI collections with expert segmentation labels and radiomic features. Nature Scientific Data 4:170117 DOI: 10.1038/sdata.2017.117. 2017.

[32] Zhou C, Ding C, Wang X, Lu Z, Tao D. One-Pass Multi-Task Networks With Cross-Task Guided Attention for Brain Tumor Segmentation. Computer Vision and Pattern Recognition. 2019.